\documentclass[10pt]{article}
\usepackage[a4paper, total={6in, 9in}]{geometry}

\usepackage[boxruled,linesnumbered,vlined,inoutnumbered]{algorithm2e}
\SetKwInOut{Parameter}{Parameters}

\usepackage{amsmath}
\usepackage{amssymb}
\usepackage{mathtools}
\usepackage{amsthm}
\usepackage{mathrsfs}
\usepackage{enumitem}
\usepackage[backend=bibtex8,
            sorting=none]{biblatex}
\usepackage{algorithm2e}

\usepackage{graphicx}
\usepackage{subfig}
\usepackage{booktabs}
\usepackage{authblk}

\DeclareMathOperator*{\pr}{\text{Pr}}
\DeclareMathOperator*{\ex}{\mathbb{E}}

\usepackage{xcolor}
\definecolor{light-grey}{rgb}{0.9,0.9,0.9}
\definecolor{dark-red}{rgb}{0.4,0.15,0.15}
\definecolor{dark-blue}{rgb}{0,0,0.7}

\usepackage{hyperref}
\hypersetup{
    colorlinks, linkcolor={dark-blue},
    citecolor={dark-blue}, urlcolor={dark-blue}
}

\title{Structural Credit Assignment with Coordinated Exploration}
\author{Stephen Chung\thanks{University of Massachusetts Amherst, Department of Computer Science, \texttt{minghaychung@umass.edu}}}
\date{April 2022} 
\begin{document}
\maketitle 
\section{Overview} 

A biologically plausible method for training an Artificial Neural Network (ANN) involves treating each unit as a stochastic Reinforcement Learning (RL) agent, thereby considering the network as a team of agents. Consequently, all units can learn via REINFORCE, a local learning rule modulated by a global reward signal, which aligns more closely with biologically observed forms of synaptic plasticity \cite[Chapter~15]{sutton2018reinforcement}. However, this learning method tends to be slow and does not scale well with the size of the network. This inefficiency arises from two factors impeding effective structural credit assignment: (i) all units independently explore the network, and (ii) a single reward is used to evaluate the actions of all units.

Accordingly, methods aimed at improving structural credit assignment can generally be classified into two categories. The first category includes algorithms that enable coordinated exploration among units, such as MAP propagation \cite{chung2021map}. The second category encompasses algorithms that compute a more specific reward signal for each unit within the network, like Weight Maximization and its variants \cite{chung2022learning}. In this research report, our focus is on the first category. We propose the use of Boltzmann machines \cite{ackley1985learning} or a recurrent network for coordinated exploration. We show that the negative phase, which is typically necessary to train Boltzmann machines, can be removed. The resulting learning rules are similar to the reward-modulated Hebbian learning rule  \cite{gerstner2014neuronal}. Experimental results demonstrate that coordinated exploration significantly exceeds independent exploration in training speed for multiple stochastic and discrete units based on REINFORCE, even surpassing straight-through estimator (STE) backpropagation \cite{bengio2013estimating}.

\subsection{An Example: How should we vote?} 

Before we begin, let us consider an intuitive example of how coordinated exploration may help structural credit assignment. Suppose that you are in a town of 99 people. At the end of each year, all people in the town vote for a mayor next year and there are always two candidates in the election: John and Peter. The candidate with more votes wins and will influence the economy of the town next year. This is the first year, and people know nothing about the two candidates. Therefore, you and the others decide whom to vote by flipping coins. Suppose you voted for John, but Peter won, and Peter performed poorly as a mayor next year. The question is—whom should you vote for next year?

REINFORCE says that you should increase doing the action that leads to a positive reward and vice versa. In this case, you should vote more for Peter the next year because you got a negative reward (the economy was poor next year) after voting for John. Though it may not seems sensible, you can still learn the optimal candidate to vote for with this learning method in the long run—when there are 49 people voting for each person (this happens with the probability of $\binom{98}{49}0.5^{49} \approx 0.0789$, which decays to $0$ quickly when there are more people in the town) and you can determine the voting result, you can learn correctly with REINFORCE; other than that, you are only learning from noise as you have no influences over the election. This illustrates why REINFORCE does not scale well with the size of networks.

MAP propagation is similar to REINFORCE, but you `forgot' the person you voted for and `believed' that you have voted for the winning candidate. In this case, all people in the town, including you, believed that they had voted for Peter instead. The negative reward is therefore associated with the action of voting Peter, and all people learn to vote more for John the next year. This is like coordination in hindsight—all people believed that they had voted for the same candidate. However, there was actually no coordination between people when deciding whom to vote, since all people decided whom to vote by flipping their own coins.

Another form of coordinated exploration is through letting people in the town communicate with one another before voting. After flipping the coin, you were prepared to vote for John. But you observed that the majority was prepared to vote for Peter, and so you followed the majority by voting for Peter. The negative reward is therefore associated with the action of voting Peter, and you learn to vote more for John the next year.

This example suggests that coordinated exploration, whether in hindsight or not, may help with structural credit assignment. We will focus on the second case, i.e.\ coordinated exploration without hindsight, in this report.

\section{Forms of Exploration}

For simplicity, we consider an MDP with only immediate reward\footnote{The algorithms in this paper can be applied in general MDPs by replacing the reward with the sum of discounted reward (i.e.\ return) or TD error, and can be applied in supervised learning tasks by replacing the reward with the negative loss.} defined by a tuple $(\mathcal{S}, \mathcal{A}, R, d_0)$, where $\mathcal{S}$ is the set of states, $\mathcal{A}$ is the set of actions, $R: \mathcal{S}\times \mathcal{A} \rightarrow \mathbb{R}$ is the reward function, and $d_0: \mathcal{S} \rightarrow [0,1]$ is the initial state distribution. Denoting the state, action, and reward by  $S$, $A$, and $R$ respectively, $\pr(S=s) = d_0(s)$ and $\ex[R|S=s, A=a] = R(s, a)$. We are interested in learning the policy $\pi:  \mathcal{S}\times \mathcal{A} \rightarrow [0,1]$ such that selecting actions according to $\pr(A=a|S=s)=\pi(s,a)$ maximizes the expected reward $\ex[R]$. 

\subsection{Independent Exploration}

Let us consider the case where the policy is computed by an ANN with one hidden layer of $N$ Bernoulli-logistic units: to sample action $A$ conditioned on $S$, we first sample the vector of activation values of the hidden layer, denoted as $H$, by:
\begin{equation}
	\pr(H=h|S=s) \propto \exp((W^Ts + b)^T h) \label{eq:exp1},
\end{equation}
where $h \in \{0, 1\}^N$, $W$ is the weight matrix of the hidden layer, and $b$ is the bias vector of the hidden layer. Equivalently, all $H_i$ (we use subscript $i$ to denote entry $i$ in a vector, and subscript $ij$ to denote entry $(i,j)$ in a matrix) are independent with each other conditioned on $S$, and the conditional distribution is given by:
\begin{equation}
	\pr(H_{i}=h|S=s) = \begin{cases}
		 1 - \sigma \left(\sum_{j} W_{ji} s_j + b_i \right), & \text{if $h = 0$,}\\
   		 \sigma \left(\sum_{j} W_{ji} s_j + b_i \right), & \text{if $h = 1$,}
				   \end{cases}			   
\end{equation}
where $h \in \{0, 1\}$ and $\sigma$ denotes the sigmoid function. After sampling $H$, we sample $A$ from an appropriate distribution conditioned on $H$. For example, if the set of actions $\mathcal{A} = \{0, 1\}$, we can use an output layer of a single Bernoulli-logistic unit:
\begin{equation}
	\pr(A=a|H=h) = \begin{cases}
		1 - \sigma \left(\sum_{j} W^{out}_{j} h_j + b^{out} \right), & \text{if $a = 0$,}\\
		\sigma \left(\sum_{j} W^{out}_{j} h_j + b^{out} \right), & \text{if $a = 1$,}
	\end{cases}	
\end{equation}
where $a \in \{0, 1\}$, $W^{out}$ is the weight vector of the output layer, and $b^{out}$ is the scalar bias of the output layer. The parameters to be learned for this network are: $W, b, W^{out}, b^{out}$.\\

\noindent \textbf{Learning Rule --} To derive a learning rule for $W$, we first compute the gradient of the expected reward $\ex[R]$ w.r.t. $W$:
\begin{align}
	\nabla_{W_{ji}} \ex[R] &= \ex[R \nabla_{W_{ji}} \log \pr(H|S) ] \\						
                        &= \ex[R (H_i - \sigma (\sum_{j} W_{ji} S_j + b_i )) S_j ], \\
                        &= \ex[R (H_i - \ex[H_i|S]) S_j]. 
\end{align}
Therefore, to perform gradient ascent on the expected reward, we can use the following REINFORCE learning rule \cite{williams1992simple}:
\begin{align}
	W_{ji} &\leftarrow W_{ji} + \alpha R (H_i - \ex[H_i|S]) S_j, \label{eq:rei1}
\end{align}
where $\alpha > 0$ denotes the step size. Also note that $\ex[H_i|S] = \sigma \left(\sum_{j} W_{ji} S_j + b_i \right)$.

The learning rule is intuitive: the expectation of $R (H_i -\ex[H_i|S])$ is simply the covariance between $R$ and $H_i$ (conditioned on $S$); that is, $\ex[R (H_i -\ex[H_i|S])|S] = Cov(R, H_i |S)$. Hence if the covariance is positive, the learning rule makes unit $i$ to fire more, and vice versa.

Recall in statistics that the covariance between two random variables $X$ and $Y$ equals:
\begin{equation}
	Cov(X, Y) = \ex[(X - \ex[X])(Y - \ex[Y])] = \ex[X(Y - \ex[Y])] = \ex[(X - \ex[X])Y],
\end{equation}
and thus the following three estimators that estimate the covariance are all unbiased: (i) $(X - \ex[X])(Y - \ex[Y])$, (ii) $X(Y - \ex[Y])$, (iii) $(X - \ex[X])Y$. We can further show that the first estimator has the lowest variance among the three. In other words, an estimator with two-sided centering gives a lower variance than an estimator with one-sided centering, and an estimator with one-sided centering is already an unbiased estimator of the covariance.

Noting that (\ref{eq:rei1}) only uses one-sided centering to estimate the covariance, an alternative learning rule with the same expected update is to use two-sided centering:
\begin{align}
	W_{ji} &\leftarrow W_{ji} + \alpha (R - \ex[R|S]) (H_i - \ex[H_i|S]) S_j, \label{eq:rei2} 
\end{align}
which is equivalent to REINFORCE with $\ex[R|S]$ as baseline \cite{williams1992simple}. $\ex[R|S]$ is generally unknown and so needs to be estimated by another network which is usually called a critic network.

Finally, we can also do one-sided centering on $R$ only, leading to the following learning rule that has the same expected update:
\begin{align}
	W_{ji} &\leftarrow W_{ji} + \alpha (R - \ex[R|S]) H_i S_j, \label{eq:rei3} 
\end{align}
which has the property that the parameter of a unit is not updated if the unit does not fire—when $H_i=0$, the above updates equal zero. This property may be desirable as biological learning rules such as reward-modulated spike-timing-dependent plasticity (R-STDP) \cite{gerstner2014neuronal} also require a neuron to fire to affect synaptic strength. 

All three learning rules (\ref{eq:rei1}), (\ref{eq:rei2}) and (\ref{eq:rei3}) follow gradient ascent on $\ex[R]$ in expectation. Among the three learning rules, learning rule (\ref{eq:rei2}) is associated with the lowest variance due to the use of two-sided centering. However, as we will see in the next section, learning rule (\ref{eq:rei3}) can be generalized more easily to incorporate coordinated exploration.

Note that a similar discussion applies to the bias $b$, and the learning rules of $b_i$ are the same as above but with $S_j$ set to $1$.

\subsection{Coordinated Exploration with Boltzmann Machines} \label{sec:bm}

As discussed previously, independent exploration coupled with a uniform reward signal makes learning inefficient and scales poorly with the number of units in the network. In this section we consider how to generalize the above network to have coordinated exploration.

Let us consider the conditional distribution of $H$ given by (\ref{eq:exp1}), which says that hidden units are conditionally independent with one another. A natural way to allow interaction between hidden units is to add a cross term between hidden units:
\begin{equation}
	\pr(H=h|S=s) \propto \exp((W^Ts + b)^T h + c h^T W^{rec} h) \label{eq:exp2},
\end{equation}
where $c \geq 0$ is a scalar hyperparameter that controls the strength of interaction between hidden units, and $W^{rec}$ is a symmetric weight matrix with a zero diagonal that determines the interaction dynamics between hidden units. We call $W^{rec}$ the recurrent weight and $W$ the feedforward weight to distinguish between these two weights. Note that we recover the case of independent exploration in (\ref{eq:exp1}) by setting $c=0$. The distribution given by (\ref{eq:exp2}) can be viewed as a fully connected Boltzmann machine \cite{ackley1985learning}, with $W^{rec}$ being the weight and $W^Ts + b$ being the bias of the Boltzmann machine. 

\begin{figure}[h!!!]
	\centering
	\includegraphics[width=0.95\textwidth]{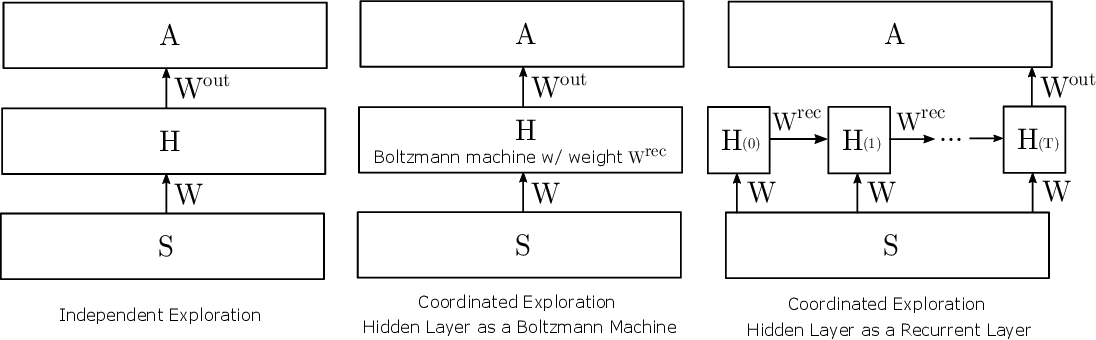}
	\caption{Illustration of how $H$ is generated by independent exploration, coordinated exploration with Boltzmann machines and recurrent networks. Left figure: a standard one-hidden-layer network of Bernoulli-logistic units. Each $H_i$ is sampled from $Ber((W^TS +b)_i)$. Middle figure: the distribution of $H$ is given by a Boltzmann machine with weight $W^{rec}$ and bias $W^TS + b$. Right figure: the hidden layer is a recurrent layer with recurrent weight $W^{rec}$ and common input $W^TS + b$, so each $H_{(t+1), i}$ is sampled from $Ber((c(W^{rec})^TH_{(t)}+W^TS +b)_i)$. If $W^{rec}$ is symmetric and has a zero diagonal, as $T \rightarrow \infty$, the distribution of $H$ depicted in the middle figure and that of $H_{(T)}$ depicted in the right figure become the same (to be rigorous, only one random entry of $H_{(t)}$ is updated in each recurrent step in the right figure for this to hold as asynchronous sampling is required).
	}
	\label{fig:0}
\end{figure}

We can then compute the distribution of $H_i$ conditioned on $H_{-i}$  and S ($H_{-i}$ denotes all entries of $H$ except $H_i$):
\begin{equation}
	\pr(H_{i}=h_i|H_{-i}=h_{-i}, S=s) = \begin{cases}
		1 - \sigma \left(c\sum_{j} W^{rec}_{ji} h_j + \sum_{j} W_{ji} s_j + b_i \right), & \text{if $h_i = 0$,}\\
		\sigma \left(c\sum_{j} W^{rec}_{ji} h_j + \sum_{j} W_{ji} s_j + b_i \right), & \text{if $h_i = 1$.}
	\end{cases} \label{eq:exp3}
\end{equation}

To sample $H$, we can perform Gibbs sampling: we first initialize $H_{(0)}$ to some values such as a zero vector (we use the subscript $(t)$ to denote time step $t$ as opposed to entry), then we randomly select one unit and replace its activation value by sampling from (\ref{eq:exp3}) to obtain $H_{(1)}$. We continue this way to obtain $H_{(2)}, H_{(3)}, ...$, and it can be shown that the limiting distribution of $H_{(t)}$ as $t \rightarrow \infty$ is the same as the distribution given by (\ref{eq:exp2}).

However, this asynchronous sampling method is computationally expensive\footnote{However, asynchronous computation should be closer to the dynamics of biological neurons as biological neurons do not fire at fixed intervals.}. Therefore, we use synchronous sampling method instead, that is, we update the activation values of all hidden units on the same time step:
\begin{equation}
H_{(t+1), i} | H_{(t)}, S \sim Ber\left(\sigma \left(c\sum_{j} W^{rec}_{ji} H_{(t), j} + \sum_{j} W_{ji} S_j + b_i \right)\right). \label{eq:ss1}
\end{equation}
This allows easy sampling of $H_{(t+1)}$ conditioned on $H_{(t)}$ and $S$. Then we let $H=H_{(T)}$, where $T \geq 1$ is a hyperparameter determining the number of steps for Gibbs sampling, and $H$ is passed to the next layer as input. See Figure \ref{fig:0} for an illustration.

Though the limiting distribution of $H_{(t)}$ as $t \rightarrow \infty$ is not guaranteed to converge due to synchronous sampling, we found that it works well in experiments. We also found in experiments that $T \geq 25$ works quite well.\\

\noindent \textbf{Learning Rule --} 
We now consider how to train a network when the distribution of $H$ is given by (\ref{eq:exp2}). Let us first compute $\nabla_{W^{rec}} \log \pr(H=h|S=s)$, which will be useful when deriving learning rules later on:
\begin{align}
&\nabla_{W^{rec}_{ji}} \log \pr(H=h|S=s) \\
=& \nabla_{W^{rec}_{ji}} \log \frac{ \exp((W^Ts + b)^T h + c h^T W^{rec} h)}{\sum_h \exp((W^Ts + b)^T h + c h^T W^{rec} h)} \\
=& c h_i h_j - \nabla_{W^{rec}_{ji}} \log \sum_h \exp((W^Ts + b)^T h + c h^T W^{rec} h) \\
=& c h_i h_j - \frac{\sum_h \exp((W^Ts + b)^T h + c h^T W^{rec} h) \nabla_{W^{rec}_{ji}} ((W^Ts + b)^T h + c h^T W^{rec} h)}{\sum_h \exp((W^Ts + b)^T h + c h^T W^{rec} h)}   \\
=& c h_i h_j - c \sum_h \pr(H=h|S=s) h_i h_j \\
=& c (h_i h_j - \ex[H_i H_j|S=s]). \label{eq:lg1}
\end{align}
This derivation is similar to the one used to derive the learning rule of Contrastive Divergence for Boltzmann machines. The term $\ex[H_i H_j|S=s]$ is called negative statistics and is usually estimated by drawing samples from the Boltzmann machine in the unclamped phase.

Similarly, we can show that:
\begin{align}
	\nabla_{W_{ji}} \log \pr(H=h|S=s) &=  c (h_i  - \ex[H_i|S=s]) s_j, \\
	\nabla_{b_{i}} \log \pr(H=h|S=s) &=  c (h_i  - \ex[H_i|S=s]).
\end{align}

Now we are ready to compute the gradient of the expected reward w.r.t. the parameters. We first consider the gradient of the expected reward w.r.t. $W^{rec}$:
\begin{align}
	\nabla_{W^{rec}_{ji}} \ex[R] &= \ex[R \nabla_{W^{rec}_{ji}} \log \pr(H|S) ] \\						
	&= \ex[c R (H_i H_j - \ex[H_i H_j|S]) ]. 
\end{align}
The last line follows from (\ref{eq:lg1}). The gradient w.r.t other parameters are similar:
\begin{align}
	\nabla_{W_{ji}} \ex[R] &= \ex[c R (H_i - \ex[H_i|S]) S_j ], \\
	\nabla_{b_{i}} \ex[R] &= \ex[c R (H_i - \ex[H_i|S]) ]. 
\end{align}
Therefore, to perform gradient ascent on the expected reward, we can use the following learning rules:
\begin{align}
	W^{rec}_{ji} &\leftarrow W^{rec}_{ji} + \alpha c R (H_i H_j - \ex[H_i H_j|S]), \label{eq:bm1} \\
	W_{ji} &\leftarrow W_{ji} + \alpha R (H_i - \ex[H_i|S]) S_j, \label{eq:bm2}	 \\
	b_{i} &\leftarrow b_{i} + \alpha R (H_i - \ex[H_i|S]). \label{eq:bm3}
\end{align}
Note that (\ref{eq:bm2}) is the same as (\ref{eq:rei1}), but the expected value $\ex[H_i|S]$ has to be computed or estimated differently since the distributions of $H$ are not the same.

These learning rules require the computation or estimation of the negative statistics $\ex[H_i H_j|S]$ and $\ex[H_i |S]$. Both terms are intractable to compute directly, and so we have to resort to Monte Carlo estimation as in Boltzmann machines: we can perform multiple Gibbs samplings to obtain samples of $H$ conditioned on $S$, and average these samples to estimate the negative statistics. But this is hardly biologically plausible and also computationally inefficient. Is it possible to remove the negative statistics from the learning rules while still performing gradient ascent?

Let us consider the terms in the learning rules. $R (H_i H_j - \ex[H_i H_j|S])$ is estimating the covariance between $R$ and $H_i H_j$, and $R(H_i - \ex[H_i|S])$ is estimating the covariance between $R$ and $H_i$. Thus, the discussion in the previous section on different estimators for covariance can also be applied here. In particular, we can do one-sided centering on $R$ instead of $H$ to obtain the following learning rules with the same expected update:
\begin{align}
	W^{rec}_{ji} &\leftarrow W^{rec}_{ji} + \alpha c (R - \ex[R|S]) H_i H_j , \label{eq:bm4} \\
	W_{ji} &\leftarrow W_{ji} + \alpha (R - \ex[R|S]) H_i S_j, \label{eq:bm5}	 \\
	b_{i} &\leftarrow b_{i} + \alpha (R - \ex[R|S]) H_i. \label{eq:bm6}
\end{align}
These learning rules follow gradient ascent on $\ex[R]$ in expectation. Again, we require a critic network to estimate $\ex[R|S]$, but this is considerably simpler than estimating  the negative statistics. In fact, $R - \ex[R|S]$ is simply TD error in a single-time-step MDP, and can be replaced with TD error in a multiple-time-step MDP, noting that TD error is also centered. In this way, we can get rid of the negative statistics in the learning rule.

It should be noted that we need at least one-side centering for learning to occur. Consider the learning rule with no centering: $W^{rec}_{ji} \leftarrow W^{rec}_{ji} + \alpha c R H_i H_j$; if the reward signals from the environment are all non-negative (e.g.\ Cartpole), then the weight can only be increased but not decreased, and learning cannot occur. 

The learning rules are similar to reward-modulated Hebbian learning \cite{gerstner2014neuronal}—for example,  (\ref{eq:bm4}) is the product between three terms: (i) TD error, (ii) the pre-synaptic activation value $H_j$ and (iii) the post-synaptic activation value $H_i$. Furthermore, as long as $W^{rec}$ is initialized to be symmetric, it will stay symmetric during training since the updates to $W^{rec}_{ji}$ and $W^{rec}_{ij}$ are the same, so we do not need additional mechanisms to ensure symmetric connections as in backprop. 

\begin{algorithm}
	\textbf{Algorithm Hyperparameter:} Number of sampling steps $T \geq 1$; strength of recurrent connections $c \geq 0$; number of hidden units $N \geq 1$; step size $\alpha > 0$\;
	\textbf{Initialize parameters:} $W \in \mathbb{R}^{dim(S) \times N}, b \in \mathbb{R}^{N}, W^{rec} \in \mathbb{R}^{N \times N}$ (symmetric with zero diagonal), parameter for the output layer $\theta^{out}$ \;
	\SetKwProg{lf}{Loop forever:}{}{} 
	\lf{}{
		Sample state $S$ \;
		\tcc{Sampling $H$ and $A$}  	 
		Sample $H_i \sim Ber\left(\sigma \left(\sum_{j} W_{ji} S_j + b_i \right)\right)$ for all $i \in \{1, 2, ..., N\}$\;
		\For{$t:=1, 2, ..., T$}{ 
			Sample $H'_i \sim Ber\left(\sigma \left(c\sum_{j} W^{rec}_{ji} H_j + \sum_{j} W_{ji} S_j + b_i \right)\right)$ for all $i \in \{1, 2, ..., N\}$ \;
			$H \leftarrow H'$ \;
		}{}		
		Sample $A \sim \pr(A=\cdot| H; \theta^{out})$ \;
		Take action $A$, observe $R$ and receive estimate $\hat{\ex}[R|S]$ from critic network\;
		\tcc{REINFORCE}  	 	
		$W^{rec}_{ji} \leftarrow W^{rec}_{ji} + \alpha c (R - \hat{\ex}[R|S]) H_i H_j$ for all $i \neq j, i, j \in \{1, 2, ..., N\}$ \;
		$W_{ji} \leftarrow W_{ji} + \alpha (R - \hat{\ex}[R|S]) H_i S_j$ for all $i \in \{1, 2, ..., N\}, j \in \{1, 2, ..., dim(S)\}$ \;
		$b_{i} \leftarrow b_{i} + \alpha (R - \hat{\ex}[R|S]) H_i$ for all $i \in \{1, 2, ..., N\}$\;
		$\theta^{out} \leftarrow \theta^{out}  + \alpha (R - \hat{\ex}[R|S]) \nabla_{\theta^{out}} \log P(A|H)$\;
		Train the critic network with error signal $R - \hat{\ex}[R|S]$\;
	}{}	
	\caption{Coordinated Exploration with a Boltzmann Machine} \label{alg:1}
\end{algorithm}

\begin{figure}[h!!!]
	\centering
	\includegraphics[width=0.95\textwidth]{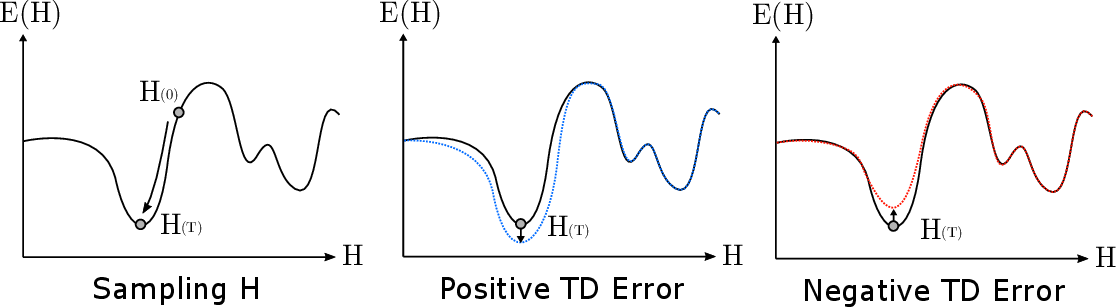}
	\caption{Illustration of learning by coordinated exploration with Boltzmann machines. First, we initialize $H_{(0)}$ and then minimize the energy, $E(H) = -\log P(H|S)$, by updating $H$ stochastically to obtain $H_{(T)}$, as shown in the left figure. Then, we pass $H_{(T)}$ to the next layer and receive the TD error. Finally, if the TD error is positive, we decrease the energy at  $H_{(T)}$ by updating the parameters of the network, as shown by the shift from the black curve to the blue dotted curve in the middle figure. This makes $H$ easier to `fall' into the same location in future sampling. The case for negative TD error is the opposite and is shown in the right figure.}
	\label{fig:1}
\end{figure}

The full algorithm is given in Algorithm \ref{alg:1}. An illustration of the algorithm from the perspective of energy-based learning is shown in Figure \ref{fig:1}. Note that if we set $c$, the strength of recurrent connections, to zero, then we recover the case of independent exploration trained with REINFORCE (one-sided centering on $R$) given by the learning rule (\ref{eq:rei3}).\\

\noindent \textbf{Experiments --} 
To test the algorithm, we consider the $k$-bit multiplexer task. The input is sampled from all possible values of a binary vector of size $k + 2^k$ with equal probability. The action set is $\{0, 1\}$, and we give a reward of $+1$ if the action equals the multiplexer's output and $-1$ otherwise. We consider $k=4$ here, so the dimension of the state space is 20 and there are $2^{20}=1048576$ possible states. We call a single interaction of the network from receiving the state to receiving the reward an episode; the series of episodes as the network's performance increases during training is called a run.

The update step size $\alpha$ for gradient ascent is chosen as $0.005$, and the batch size is chosen as $16$ (i.e.\ we estimated the gradient for $16$ episodes and averaged them in the update). We used the Adam optimizer \cite{kingma2014adam}. These hyperparameters are chosen prior to the experiments without any tuning. We use a one-hidden-layer ANN as the critic network and train it by backprop. These settings are the same for all our experiments.

\begin{figure}[h!!!]
	\centering
	\includegraphics[width=\textwidth]{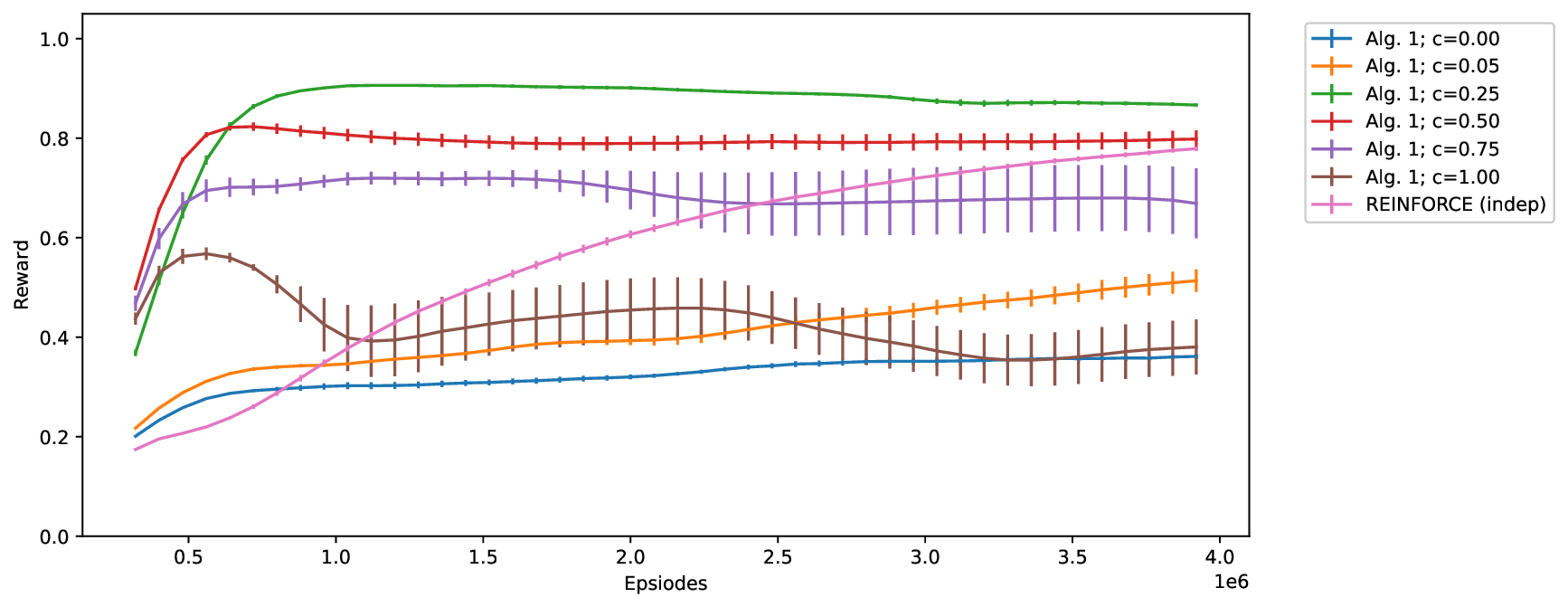}
	\caption{Learning curve of Algorithm \ref{alg:1} with varying $c$ and $T=25$. Results are averaged over 5 independent runs, and the error bar represents the standard deviation over the runs.}
	\label{fig:2}
\end{figure}

First, we tested Algorithm \ref{alg:1} with $N=64$ hidden units and $T=25$ sampling steps. The learning curve (i.e. the reward of each episode throughout a run) with varying $c$ (the strength of recurrent connection) is shown in Figure \ref{fig:2}. The figure also shows the learning curve of a network with one hidden layer of $64$ Bernoulli-logistic units trained by REINFORCE with baseline (\ref{eq:rei2}), which we label `REINFORCE (indep)'. We observe that:

\begin{itemize}
	\item Both `REINFORCE (indep)' and `Alg. 1; c=0.00' use independent exploration. The former one (\ref{eq:rei2}) use centering on both side while the latter one (\ref{eq:rei3}) only use centering on $R$. As expected, two-sided centering gives a lower variance and so the former one performs better.
	\item However, as $c$ increases from $0$ to $0.25$, the performance of Algorithm \ref{alg:1} improves due to the coordinated exploration, offsetting the reduced performance from one-sided centering.
	\item The optimal $c$ is $0.25$; at $c=0.25$, the learning curve, especially during the first $1e6$ episodes, is significantly better than `REINFORCE (indep)', indicating that Algorithm \ref{alg:1} can improve learning speed significantly with a proper $c$.
	\item But as $c$ increases larger than $0.25$, performance deteriorates since inputs from other hidden units instead of the state dominate the unit. In particular, the asymptotic performance is unstable for $c \geq 0.75$. 
\end{itemize}

The experiment result suggests that coordinated exploration with Boltzmann machines can improve learning speed; the benefit from coordinated exploration offsets the reduced performance from one-sided centering. Also, $c$, the strength of recurrent connection, is an important hyperparameter that needs to be carefully tuned. Simply setting $c=1$ can lead to reduced performance.

\begin{figure}[h!!!]
	\centering
	\includegraphics[width=\textwidth]{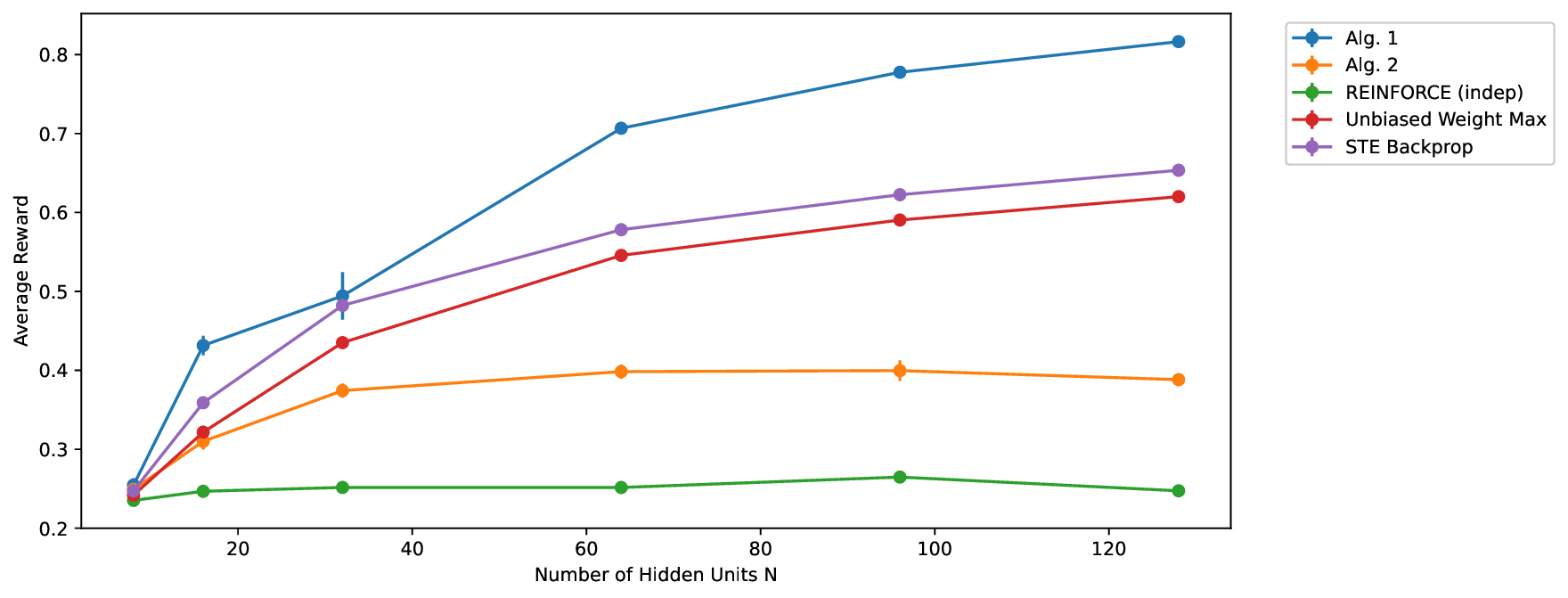}
	\caption{Average reward during the first 1e6 episodes for a network with varying $N$. Results are averaged over 5 independent runs, and the error bar represents the standard deviation over the runs.}
	\label{fig:3}
\end{figure}

\begin{figure}[h!!!]
	\centering
	\includegraphics[width=\textwidth]{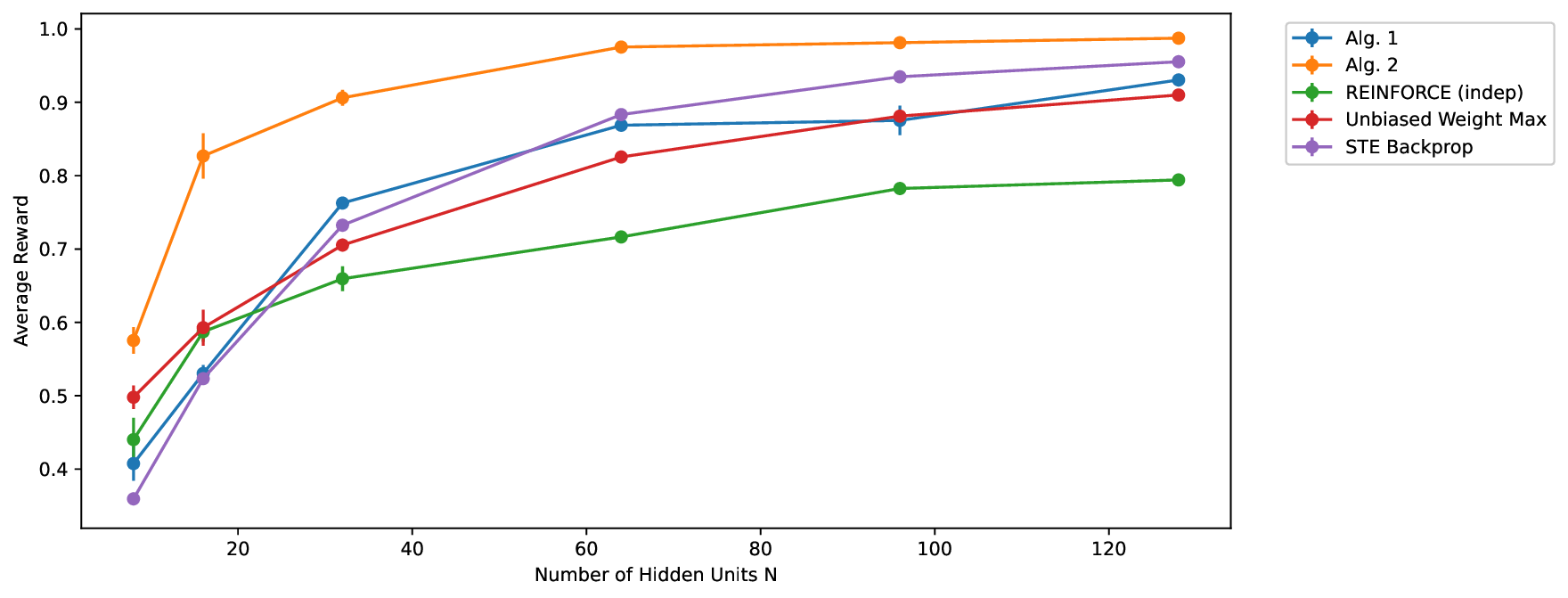}
	\caption{Average reward during the last 1e6 episodes for a network with varying $N$. Results are averaged over 5 independent runs, and the error bar represents the standard deviation over the runs.}
	\label{fig:4}
\end{figure}

To understand how the different learning rules and networks scale with $N$, the number of hidden units, we repeat the above experiments for $N \in \{8, 16, 32, 64, 96, 128\}$. We used $T=25$, and $c$ was tuned for each $N$ to maximize the average reward throughout all episodes. The average reward during the first 1e6 episodes (to evaluate the learning speed) and the last 1e6 episodes (to evaluate the asymptotic performance) for different $N$ is shown in Figure \ref{fig:3} and \ref{fig:4} respectively. For comparison, we also show the results from training a network with one hidden layer of $N$ Bernoulli-logistic units using REINFORCE with baseline (\ref{eq:rei2}), unbiased Weight Maximization, and STE backprop. 

From Figure \ref{fig:3}, we observe that Algorithm \ref{alg:1} has the best learning speed among all algorithms considered, and also scales well with the number of hidden units. It is worth noting that \emph{Algorithm \ref{alg:1} is faster than STE backprop} at all $N$ tested. From Figure \ref{fig:4}, we observe that the asymptotic performance of Algorithm \ref{alg:1} is similar to other algorithms considered except Algorithm \ref{alg:2}, another algorithm that will be discussed in the next section. However, it should be noted that the computation cost of Algorithm \ref{alg:1} is more expensive than other algorithms considered. As we use $T=25$, it means that we require $25$ times computational cost to sample $H$ compared to other algorithms.

Additional experiments, such as ablation analysis on $T$ and how the optimal $c$ changes with $N$, can be found in Appendix \ref{sec:a11}.

\subsection{Coordinated Exploration with Recurrent Networks}

Though symmetric recurrent weights arise naturally from Algorithm \ref{alg:1}, symmetric connections between biological neurons are not commonly observed. Learning rules such as R-STDP is asymmetric instead of symmetric: if the post-synaptic spike arrives before the pre-synaptic spike arrives, then the synaptic strength is decreased instead of increased. Moreover, the use of one-sided centering increases variance and hurts performance. Is it possible to remove the requirement of symmetric connection in Algorithm \ref{alg:1} while using two-sided centering?

Let look at how $H$ is generated in Algorithm \ref{alg:1}, as given by (\ref{eq:ss1}). One can also see this equation as a recurrent neural network, with $S$ being the common input at all time steps, as shown in Figure \ref{fig:0}. From this perspective, we can simply apply REINFORCE on each time step to obtain an unbiased estimator of the gradient. For example, the gradient of the expected reward w.r.t. $W^{rec}$ can be computed as:
\begin{align}
	\nabla_{W^{rec}_{ji}} \ex[R] &= \ex \left[R \sum_{t=1}^T \nabla_{W^{rec}_{ji}} \log \pr(H_{(t+1)}|S, H_{(t)}) \right] \\						
		&= \ex \left[cR \sum_{t=1}^T  (H_{(t+1), i} - \ex[H_{(t+1), i}|S, H_{(t)}]) H_{(t),j} \right],
\end{align}
where $\ex[H_{(t+1), i}|S, H_{(t)}] = \sigma \left(c\sum_{j} W^{rec}_{ji} H_{(t), j} + \sum_{j} W_{ji} S_j + b_i \right)$. The first line follows by the fact that a recurrent ANN can be viewed as a multi-layer ANN with shared weights, and so we can apply REINFORCE on each layer.

The gradient w.r.t. $W$ and $b$ can be computed similarly:
\begin{align}
	\nabla_{W_{ji}} \ex[R] &= \ex \left[R \sum_{t=1}^T  (H_{(t+1), i} - \ex[H_{(t+1), i}|S, H_{(t)}]) S_j \right], \\
	\nabla_{b_{i}} \ex[R] &= \ex \left[R \sum_{t=1}^T  (H_{(t+1), i} - \ex[H_{(t+1), i}|S, H_{(t)}]) \right].	
\end{align}
Note that these equations do not assume $W^{rec}$ to be symmetric or to have zero diagonal. It also does not require $H^{(T)}$ to converge to any limiting distribution, so $T$ can be small. From these equations, we arrive at the following learning rules that follow gradient ascent on expected reward in expectation:
\begin{align}
	W^{rec}_{ji} &\leftarrow W^{rec}_{ji} + \alpha c R \sum_{t=1}^T  (H_{(t+1), i} - \ex[H_{(t+1), i}|S, H_{(t)}]) H_{(t),j} , \label{eq:rn1} \\
	W_{ji} &\leftarrow W_{ji} + \alpha R \sum_{t=1}^T  (H_{(t+1), i} - \ex[H_{(t+1), i}|S, H_{(t)}]) S_j, \label{eq:rn2}	 \\
	b_{i} &\leftarrow b_{i} + \alpha R \sum_{t=1}^T  (H_{(t+1), i} - \ex[H_{(t+1), i}|S, H_{(t)}]). \label{eq:rn3}
\end{align}
However, these learning rules suffer from a large variance as they can be viewed as REINFORCE applied to a multi-layer ANN with shared weights. To solve this issue, we borrow the idea of temporal credit assignment in RL. Consider that we use a very large $T$, e.g.\ $T=1000$. It is unlikely that $H_{(t)}$ for a small $t$ can have influences on $H_{(1000)}$ and also the reward $R$. In other words, the dependency between $R$ and $H_{(t)}$ decreases as $t$ decreases from $T$. Thus, instead of placing equal importance on every time step in the learning rule, we can multiply the terms in the learning rule with an exponentially decaying rate $0 \leq \lambda \leq 1$:
\begin{align}
	W^{rec}_{ji} &\leftarrow W^{rec}_{ji} + \alpha c R \sum_{t=1}^T  \lambda^{T-t} (H_{(t+1), i} - \ex[H_{(t+1), i}|S, H_{(t)}]) H_{(t),j} , \label{eq:rn4} \\
	W_{ji} &\leftarrow W_{ji} + \alpha R \sum_{t=1}^T  \lambda^{T-t} (H_{(t+1), i} - \ex[H_{(t+1), i}|S, H_{(t)}]) S_j, \label{eq:rn5}	 \\
	b_{i} &\leftarrow b_{i} + \alpha R \sum_{t=1}^T  \lambda^{T-t} (H_{(t+1), i} - \ex[H_{(t+1), i}|S, H_{(t)}]). \label{eq:rn6}
\end{align}
These learning rules can be easily implemented online by eligibility traces. For example, the learning rule for $W^{rec}$ is equivalent to:
\begin{align}
	\mathbf{z}^{rec}_{(0), ji} &= 0.\\
	\mathbf{z}^{rec}_{(t+1), ji} &= \lambda \mathbf{z}^{rec}_{(t), ji} + (H_{(t+1), i} - \ex[H_{(t+1), i}|S, H_{(t)}]) H_{(t),j}, \text{for } t \in \{1, 2, ..., T\},\\
	W^{rec}_{ji} &\leftarrow \alpha c R \mathbf{z}^{rec}_{(T), ji}.
\end{align}

\begin{algorithm}
	\textbf{Algorithm Hyperparameter:} Number of sampling steps $T \geq 1$; strength of recurrent connections $c \geq 0$; number of hidden units $N \geq 1$; eligibility decay rate $0 \leq \lambda \leq 1$; step size $\alpha > 0$\;
	\textbf{Initialize parameters:} $W \in \mathbb{R}^{dim(S) \times N}, b \in \mathbb{R}^{N}, W^{rec} \in \mathbb{R}^{N \times N}$, parameter for the output layer $\theta^{out}$ \;
	\SetKwProg{lf}{Loop forever:}{}{} 
	\lf{}{
		Sample state $S$ \;
		\tcc{Sampling $H$ and $A$}  
		Sample $H_i \sim Ber\left(\sigma \left(\sum_{j} W_{ji} S_j + b_i \right)\right)$ for all $i \in \{1, 2, ..., N\}$\;	 
		Initialize $\mathbf{z}^{rec}=\mathbf{0}, \mathbf{z}=\mathbf{0}$\;
		\For{$t:=1, 2, ..., T$}{ 
			$p_i \leftarrow \sigma \left(c\sum_{j} W^{rec}_{ji} H_j + \sum_{j} W_{ji} S_j + b_i \right)$  for all $i \in \{1, 2, ..., N\}$ \;
			Sample $H'_i \sim Ber(p_i)$ for all $i \in \{1, 2, ..., N\}$ \;
			$\mathbf{z}^{rec}_{ji} \leftarrow \lambda \mathbf{z}^{rec}_{ji} + (H'_i - p_i) H_j$ for all $i, j \in \{1, 2, ..., N\}$ \;
			$\mathbf{z}_{i} \leftarrow \lambda \mathbf{z}_{i} + (H'_i - p_i)$ for all $i \in \{1, 2, ..., N\}$\;
			$H \leftarrow H'$ \;
		}{}		
		Sample $A \sim \pr(A=\cdot| H; \theta^{out})$ \;
		Take action $A$, observe $R$ and receive estimate $\hat{\ex}[R|S]$ from critic network\;
		\tcc{REINFORCE}  	 	
		$W^{rec}_{ji} \leftarrow W^{rec}_{ji} + \alpha c (R - \hat{\ex}[R|S]) \mathbf{z}^{rec}_{ji}$ for all $i, j \in \{1, 2, ..., N\}$ \;
		$W_{ji} \leftarrow W_{ji} + \alpha (R - \hat{\ex}[R|S]) \mathbf{z}_{i} S_j $ for all $i \in \{1, 2, ..., N\}, j \in \{1, 2, ..., dim(S)\}$ \;
		$b_{i} \leftarrow b_{i} + \alpha (R - \hat{\ex}[R|S])  \mathbf{z}_{i}$ for all $i \in \{1, 2, ..., N\}$\;
		$\theta^{out} \leftarrow \theta^{out}  + \alpha (R - \hat{\ex}[R|S]) \nabla_{\theta^{out}} \log P(A|H)$\;
		Train the critic network with error signal $R - \hat{\ex}[R|S]$\;
	}{}	
	\caption{Coordinated Exploration with a Recurrent Network} \label{alg:2}
\end{algorithm}

The full algorithm is given in Algorithm \ref{alg:2}. We also adjust the reward by the expected reward from a critic network in the algorithm (though this is not necessary for learning to occur, as opposed to Algorithm \ref{alg:1}). It should be noted that at $c=0$ and $\lambda=0$, the algorithm recovers REINFORCE with baseline (\ref{eq:rei2}) applied on a network with one hidden layer of $N$ Bernoulli-logistic units. 

In addition, at $\lambda=0$ (which works well in our experiments), that is, the hidden units only learn at the last time step, Algorithm \ref{alg:2} becomes the same as Algorithm \ref{alg:1}, except: (i) Algorithm \ref{alg:2} uses two-sided centering as $\ex[H_{(T)}|S, H_{(T-1)}]$ is subtracted from $H_{(T)}$, which makes the learning rule no longer symmetric and so is the recurrent weight $W^{rec}$; (ii) the pre-synaptic signal in the learning rule of the recurrent weight $W^{rec}$ is $H_{(T-1)}$ in Algorithm \ref{alg:2} instead of $H_{(T)}$ in Algorithm \ref{alg:1}. In other words, the pre-synaptic signal is one time step before the post-synaptic signal in the learning rule of $W^{rec}$ in Algorithm \ref{alg:2}, making it closer to R-STDP instead of reward-modulated Hebbian learning.

Though Algorithm \ref{alg:2} may seem similar to Algorithm \ref{alg:1} as $H$ is generated in the exact same way, the dynamics of these two algorithms are very different throughout learning. Algorithm \ref{alg:2} treats the hidden layer as a recurrent layer instead of a Boltzmann machine, so any $T \geq 1$ already gives a good performance instead of $T \geq 25$ in Algorithm \ref{alg:1}. The asymmetric connection also makes $H$ not guaranteed to converge to any distributions as in Boltzmann machines. These different dynamics manifest in our experiments, which will be discussed in the following section. \\

\noindent \textbf{Experiments --} 
To test the algorithm, we use the same setting as in Section \ref{sec:bm}; that is, we consider the task of $4$-bit multiplexer tasks with the same set of hyperparameters. First, we tested Algorithm \ref{alg:2} with $N=64$ hidden units, $T=2$ sampling steps and $\lambda=0.25$ trace decay rate. The learning curve with varying $c$ (the strength of recurrent connection) is shown in Figure \ref{fig:5}. The figure also shows the learning curve of a network with one hidden layer of $64$ Bernoulli-logistic units trained by REINFORCE with baseline (\ref{eq:rei2}), which we label `REINFORCE (indep)'. The learning curve of Algorithm \ref{alg:1} with $c=0.25$ from Figure \ref{fig:2} is also shown here for comparison. We observe that:

\begin{figure}[h!!!]
	\centering
	\includegraphics[width=\textwidth]{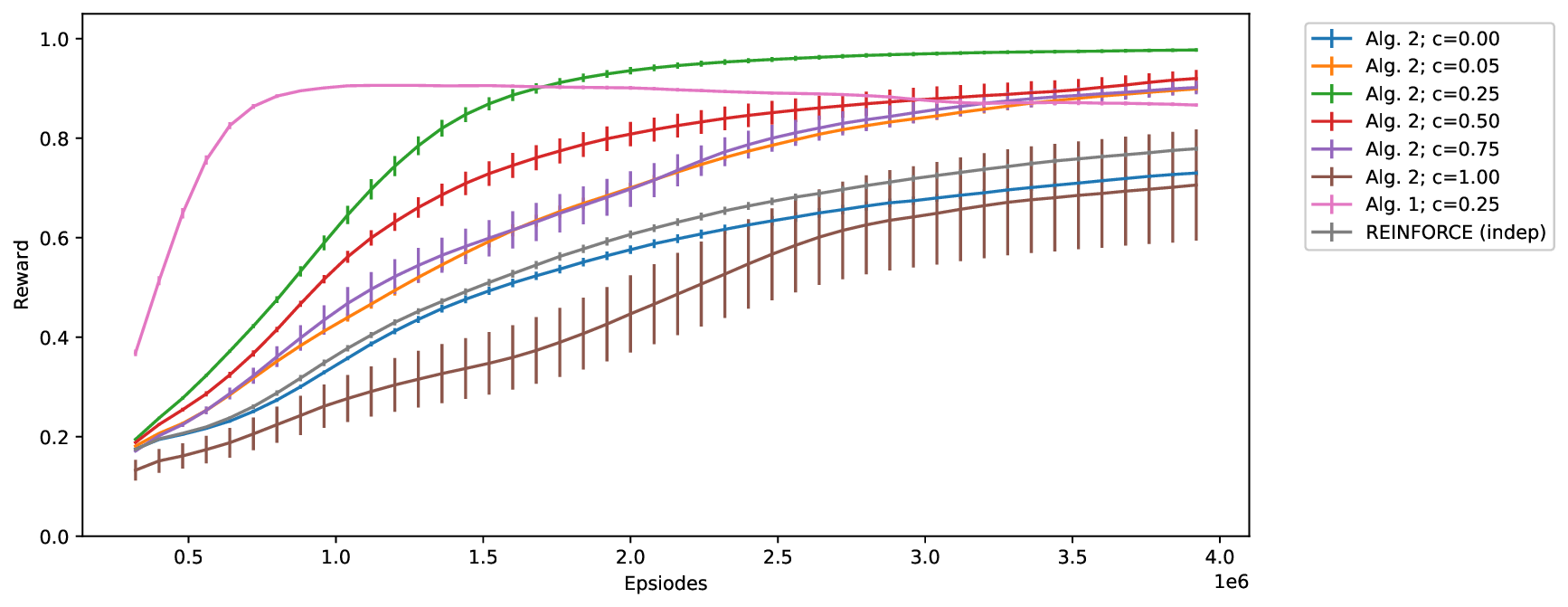}
	\caption{Learning curve of Algorithm \ref{alg:2} with varying $c$, $T=2$ and $\lambda=0.25$. Results are averaged over 5 independent runs, and the error bar represents the standard deviation over the runs.}
	\label{fig:5}
\end{figure}

\begin{itemize}
	\item  Both `REINFORCE (indep)' and `Alg. 2; c=0.00' use independent exploration, and the former one is equivalent to Algorithm \ref{alg:2} with both $c=0$ and $\lambda=0$. Since `Alg. 2; c=0.00' uses $\lambda=0.25$ and there are no interaction between hidden units, the use of $\lambda$ only adds noise to the parameter update and leads to a slightly worse performance.
	\item However, as $c$ increases from $0$ to $0.25$, the performance of Algorithm \ref{alg:2} improves due to the coordinated exploration.
	\item The optimal $c$ is $0.25$; at $c=0.25$, the learning curve is better than `REINFORCE (indep)' in terms of both learning speed and asymptotic performance; compared to `Alg. 1; c=0.25', the learning speed is lower, but the asymptotic performance is better.
	\item But as $c$ increases larger than $0.25$, performance deteriorates since inputs from other hidden units instead of the state dominate the unit.
	\item The asymptotic performance is stable for all $c$, unlike Algorithm \ref{alg:1}.
\end{itemize}
The experiment result suggests that coordinated exploration with recurrent networks can improve learning speed. Again, $c$, the strength of recurrent connection, is an important hyperparameter that needs to be carefully tuned. The dynamics of Algorithm \ref{alg:1} and \ref{alg:2} are also different; Algorithm \ref{alg:1} has a high learning speed while Algorithm \ref{alg:2} has a better and stable asymptotic performance. In addition, Algorithm \ref{alg:1} requires $T \geq 25$ for good performance while the performance of Algorithm \ref{alg:2} remains stable for all $T \geq 1$.

To understand how the different learning rules and networks scale with $N$, the number of hidden units, we repeat the above experiments for $N \in \{8, 16, 32, 64, 96, 128\}$. We used $T=2$, and $c$ was tuned for each $N$ to maximize the average reward throughout all episodes. The average reward during the first 1e6 episodes (to evaluate the learning speed) and the last 1e6 episodes (to evaluate the asymptotic performance) for different $N$ is shown in Figure \ref{fig:3} and \ref{fig:4} respectively. 

From Figure \ref{fig:3}, we observe that the learning speed of Algorithm \ref{alg:2} is moderately higher than `REINFORCE (indep)' and also does not benefit much from increasing the network's size. From Figure \ref{fig:4}, however, we observe that the asymptotic performance of Algorithm \ref{alg:2} is the best among all algorithms considered at all $N$.

The lower learning speed but better asymptotic performance can be explained by the assumption used in deriving the learning rules. In Algorithm \ref{alg:2}, we view $H$ as the output of a recurrent layer, and we apply REINFORCE to the hidden units on the last few steps (due to eligibility traces). This is equivalent to training an ANN of $T$ hidden layers with shared weight $W^{rec}$ by applying REINFORCE on the last few layers. Thus, the low learning speed of REINFORCE in training an ANN also manifests here; however, the deeper network increases the network's capacity, thereby giving a better asymptotic performance.

Additional experiments, such as ablation analysis on $T$ and $\lambda$, and how the optimal $c$ changes with $N$, can be found in Appendix \ref{sec:a12}.

\subsection{Discussion}

In this report we propose and discuss two different algorithms to allow more efficient structural credit assignment by coordinated exploration. The first algorithm generalizes hidden layers of Bernoulli-logistic units to Boltzmann machines, while the second algorithm generalizes hidden layers to recurrent layers. It turns out that the activation values of hidden layers in these two algorithms can be sampled in the exact same way. However, the resulting learning rules are different and lead to different dynamics. Coordinated exploration with Boltzmann machines (Algorithm \ref{alg:1}) increases learning speed significantly such that the learning speed is even higher than STE backprop, whereas coordinated exploration with recurrent networks (Algorithm \ref{alg:2}) improves asymptotic performance. It remains to be seen whether we can combine the advantages of both algorithms in a single learning rule. Nonetheless, in either algorithm, we observe an improved learning speed compared to the baseline of training a network of Bernoulli-logistic units by REINFORCE, indicating that coordinated exploration can facilitate structural credit assignment. 

In addition, coordinated exploration by Boltzmann machines or recurrent networks is arguably much more biologically plausible than backprop. Compared to a network of Bernoulli-logistic units trained by REINFORCE, which is argued to be close to biological learning \cite[Chapter~15]{sutton2018reinforcement}, coordinated exploration only requires adding recurrent connections between units on the same layer.

However, we only consider a one-hidden-layer ANN in this research report. Since the coordinated exploration of both Algorithm \ref{alg:1} and \ref{alg:2} is intra-layer instead of inter-layer, we expect that additional work is required to match the speed of STE backprop when applied in a multi-layer ANN. It may be possible to design a reward signal specific to each layer, with each layer being a Boltzmann machine or a recurrent layer that receives inputs from the previous layer. In this way, we have a vector reward signal with each reward signal targeting a subset of neurons, and the subset of neurons communicate with one another to generate a collective action such that a single reward is sufficient for assigning credit to the subset; this may be closer to the vector reward prediction errors observed in biological systems \cite{parker2016reward, lee2022vector}. 

Finally, further work can be done to understand how the long-term depression (LTD) in R-STDP can benefit learning, which is not included in the learning rules of both Algorithm \ref{alg:1} and \ref{alg:2}. R-STDP states that if the pre-synaptic spike arrives after the post-synaptic spike, the synaptic strength is decreased instead of increased. This rule can be incorporated directly in the learning rules of Algorithm \ref{alg:1} or \ref{alg:2}, but it remains to be seen whether it can benefit learning and what the theoretical basis is. Most studies on how R-STDP induces learning behavior \cite{florian2007reinforcement, fremaux2013reinforcement} focus on long-term potentiation (LTP) instead of LTD, and so how LTD can help learning is an important area for further work.

\section{Acknowledgment}

We would like to thank Andrew G. Barto, who inspired this research and provided valuable insights
and comments.

\clearpage
\appendix
\section{Additional Experiments}
\subsection{Coordinated Exploration with Boltzmann Machines} \label{sec:a11}

\begin{figure}[h!!!]
	\centering
	\includegraphics[width=\textwidth]{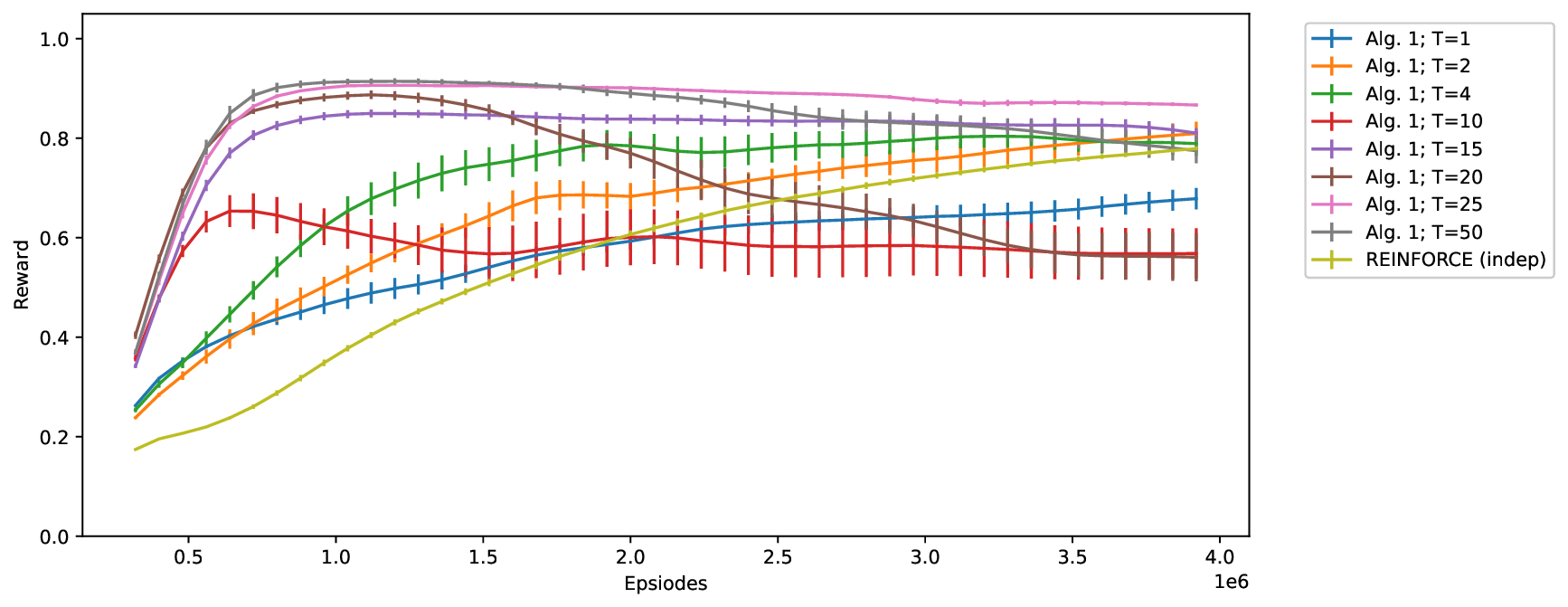}
	\caption{Learning curve of Algorithm \ref{alg:1} with varying $T$ and $c=0.25$. Results are averaged over 5 independent runs, and the error bar represents the standard deviation over the runs.}
	\label{fig:a1}
\end{figure}

To see how the performance of Algorithm \ref{alg:1} scales with $T$, the number of sampling step, we repeat the experiments shown in Figure \ref{fig:2} but with $c=0.25$ and $T \in \{1, 2, 4, 10, 15, 20, 25, 30\}$. The results are shown in Figure \ref{fig:a1}. We observe that in terms of learning speed, $T \geq 15$ works well; in terms of asymptotic performance, $T=25$ works the best. It is interesting to observe that the asymptotic performance of $T=50$ is slightly worse than $T=25$. The reason for this remains to be understood.

\begin{figure}[h!!!]
	\centering
	\includegraphics[width=0.9\textwidth]{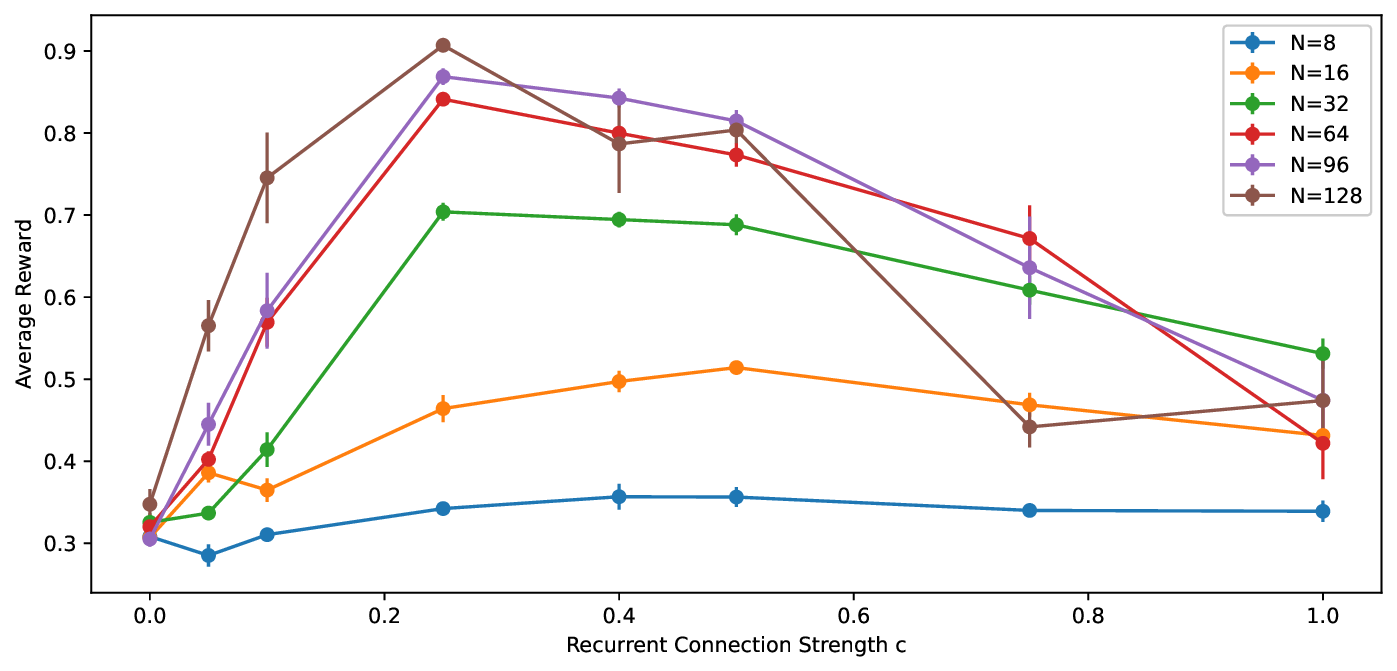}
	\caption{Average reward throughout all episodes of Algorithm \ref{alg:1} with varying $N$ and $c$. Results are averaged over 5 independent runs, and the error bar represents the standard deviation over the runs.}
	\label{fig:a2}
\end{figure}

Finally, we consider how the optimal strength of recurrent connection $c$ changes with number of hidden units $N$. The average reward throughout all episodes for Algorithm \ref{alg:1} with $N \in \{8, 16, 32, 64, 96, 128\}$ and $c \in \{0.00, 0.05, 0.10, 0.25, 0.40, 0.50, 0.75, 1.00\}$ is shown in Figure \ref{fig:a2}. We used $T=25$ here. We observe that the optimal $c$ scales inversely with $N$. The optimal $c$ when $N \leq 16$ is around $0.50$ while the optimal $c$ when $N > 16$ is around $0.25$.

\subsection{Coordinated Exploration with Recurrent Networks} \label{sec:a12}

\begin{figure}[h!!!]
	\centering
	\includegraphics[width=\textwidth]{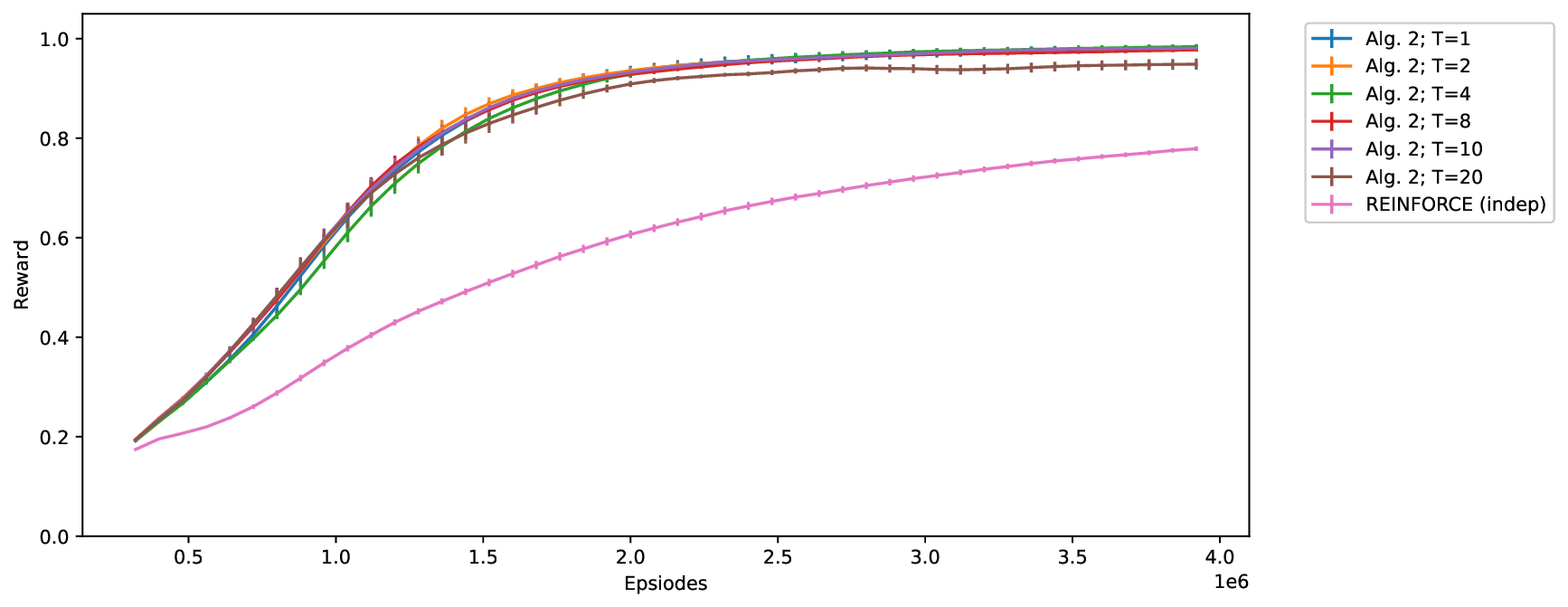}
	\caption{Learning curve of Algorithm \ref{alg:2} with varying $T$ and $c=0.25$. Results are averaged over 5 independent runs, and the error bar represents the standard deviation over the runs.}
	\label{fig:a3}
\end{figure}

To see how the performance of Algorithm \ref{alg:2} scales with $T$, the number of sampling step, we repeat the experiments shown in Figure \ref{fig:5} but with $c=0.25$ and $T \in \{1, 2, 4, 8, 10, 20\}$. The results are shown in Figure \ref{fig:a3}. We observe that there are almost no differences in the learning curves for different $T$.

\begin{figure}[h!!!]
	\centering
	\includegraphics[width=\textwidth]{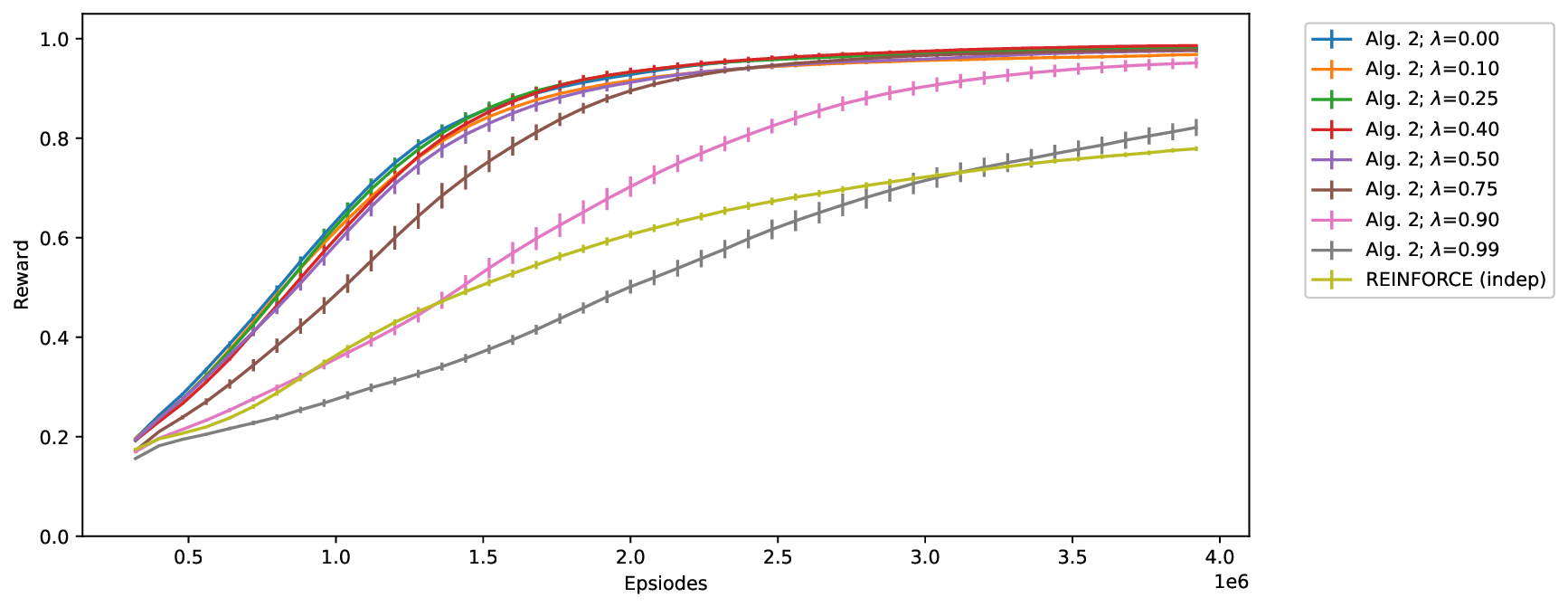}
	\caption{Learning curve of Algorithm \ref{alg:2} with varying $\lambda$ and $c=0.25$. Results are averaged over 5 independent runs, and the error bar represents the standard deviation over the runs.}
	\label{fig:a4}
\end{figure}

To see how the performance of Algorithm \ref{alg:2} scales with $\lambda$, the trace decay rate, we repeat the experiments shown in Figure \ref{fig:5} but with $c=0.25$ and $\lambda \in \{0.00, 0.10, 0.25, 0.40, 0.50, 0.75, 0.90, 0.99\}$. The results are shown in Figure \ref{fig:a4}. We observe that there are almost no differences in the learning curves for $\lambda \leq 0.50$, indicating that we can apply REINFORCE on the last time step and the performance is still the same. However, performance deteriorates as $\lambda$ grows larger than $0.50$ due to the increased noise.

\begin{figure}[h!!!]
	\centering
	\includegraphics[width=0.9\textwidth]{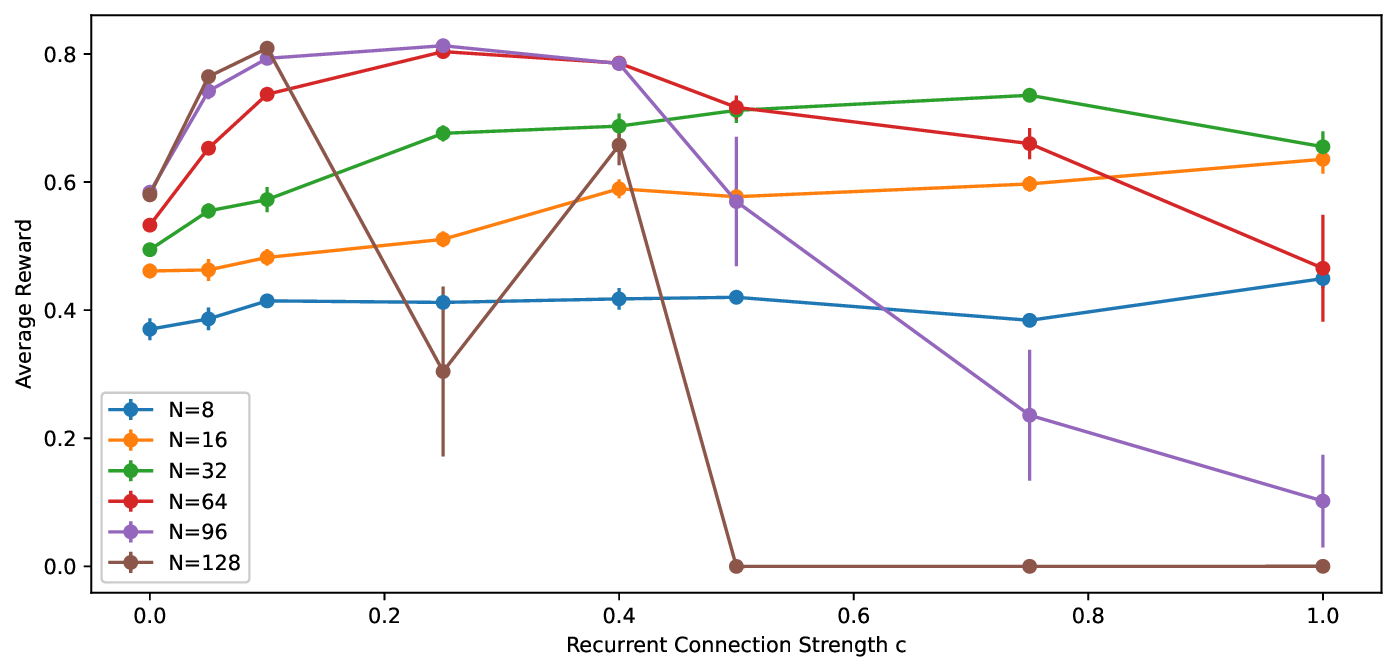}
	\caption{Average reward throughout all episodes of Algorithm \ref{alg:2} with varying $N$ and $c$. Results are averaged over 5 independent runs, and the error bar represents the standard deviation over the runs.}
	\label{fig:a5}
\end{figure}

Finally, we consider how the optimal strength of recurrent connection $c$ changes with the number of hidden units $N$. The average reward throughout all episodes for Algorithm \ref{alg:2} with $N \in \{8, 16, 32, 64, 96, 128\}$ and $c \in \{0.00, 0.05, 0.10, 0.25, 0.40, 0.50, 0.75, 1.00\}$ is shown in Figure \ref{fig:a5}. We used $T=2$ and $\lambda=0.25$ here. We observe that the optimal $c$ scales inversely with $N$. The optimal $c$ for $N= 8, 16, 32, 64, 96, 128$ are respectively $1.00, 1.00, 0.75, 0.25, 0.25, 0.10$.

\subsection{Miscellaneous} \label{sec:a13}

\begin{figure}[h!!!]
	\centering
	\includegraphics[width=\textwidth]{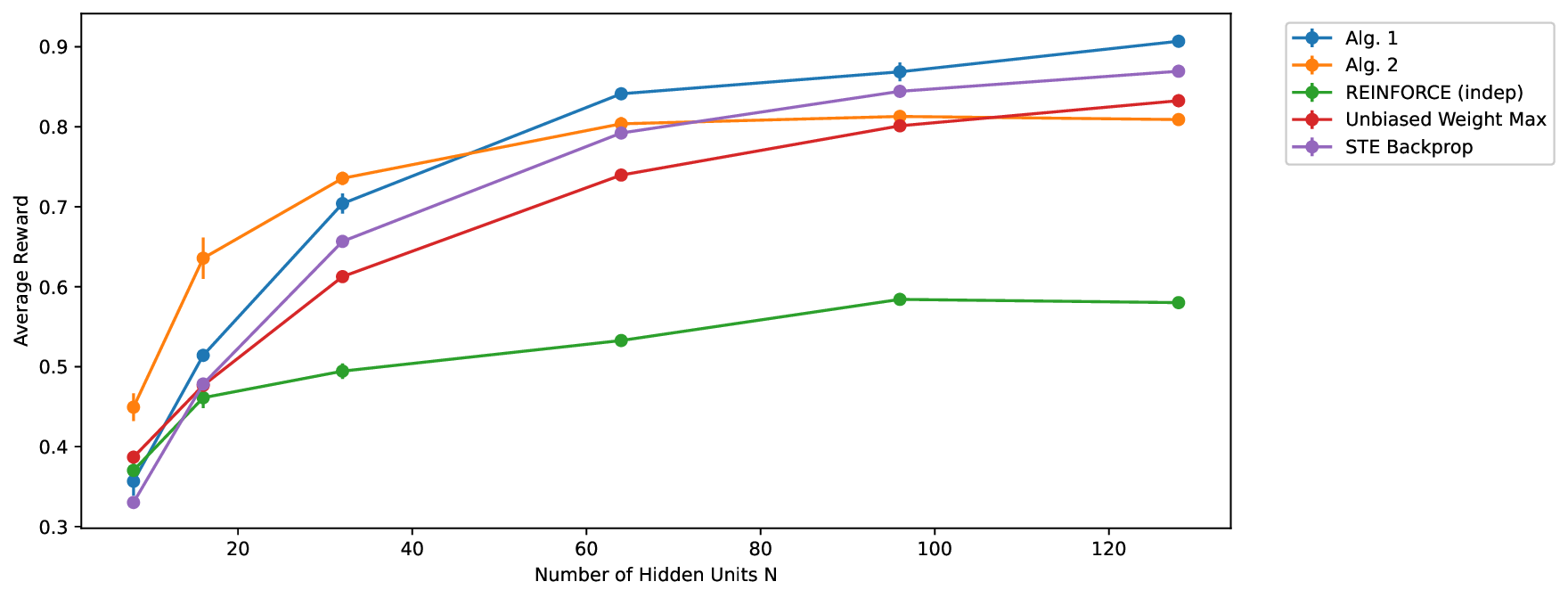}
	\caption{Average reward throughout all episodes for a network with varying $N$. Results are averaged over 5 independent runs, and the error bar represents the standard deviation over the runs.}
	\label{fig:a6}
\end{figure}

For completeness, we repeat Figure \ref{fig:3} and \ref{fig:4} here, but we show the average reward throughout all 4e6 episodes instead of only the first or the last 1e6 episodes. The results are shown in Figure \ref{fig:a6}.

\clearpage
\printbibliography

@article{williams1992simple,
  title={Simple statistical gradient-following algorithms for connectionist reinforcement learning},
  author={Williams, Ronald J},
  journal={Machine learning},
  volume={8},
  number={3-4},
  pages={229--256},
  year={1992},
  publisher={Springer}
}

@article{bengio2013estimating,
  title={Estimating or propagating gradients through stochastic neurons for conditional computation},
  author={Bengio, Yoshua and L{\'e}onard, Nicholas and Courville, Aaron},
  journal={arXiv preprint arXiv:1308.3432},
  year={2013}
}

@article{kingma2014adam,
  title={Adam: A method for stochastic optimization},
  author={Kingma, Diederik P and Ba, Jimmy},
  journal={arXiv preprint arXiv:1412.6980},
  year={2014}
}

@book{gerstner2014neuronal,
  title={Neuronal dynamics: From single neurons to networks and models of cognition},
  author={Gerstner, Wulfram and Kistler, Werner M and Naud, Richard and Paninski, Liam},
  year={2014},
  publisher={Cambridge University Press}
}

@inproceedings{chung2022learning,
	title={Learning by competition of self-interested reinforcement learning agents},
	author={Chung, Stephen},
	booktitle={Proceedings of the AAAI Conference on Artificial Intelligence},
	volume={36},
	number={6},
	pages={6384--6393},
	year={2022}
}

@article{chung2021map,
	title={MAP Propagation Algorithm: Faster Learning with a Team of Reinforcement Learning Agents},
	author={Chung, Stephen},
	journal={Advances in Neural Information Processing Systems},
	volume={34},
	year={2021}
}

@article{ackley1985learning,
	title={A learning algorithm for Boltzmann machines},
	author={Ackley, David H and Hinton, Geoffrey E and Sejnowski, Terrence J},
	journal={Cognitive science},
	volume={9},
	number={1},
	pages={147--169},
	year={1985},
	publisher={Elsevier}
}

@article{parker2016reward,
	title={Reward and choice encoding in terminals of midbrain dopamine neurons depends on striatal target},
	author={Parker, Nathan F and Cameron, Courtney M and Taliaferro, Joshua P and Lee, Junuk and Choi, Jung Yoon and Davidson, Thomas J and Daw, Nathaniel D and Witten, Ilana B},
	journal={Nature neuroscience},
	volume={19},
	number={6},
	pages={845--854},
	year={2016},
	publisher={Nature Publishing Group}
}

@article{lee2022vector,
	title={A vector reward prediction error model explains dopaminergic heterogeneity},
	author={Lee, Rachel S and Engelhard, Ben and Witten, Ilana B and Daw, Nathaniel D},
	journal={bioRxiv},
	year={2022},
	publisher={Cold Spring Harbor Laboratory}
}

@book{sutton2018reinforcement,
	title={Reinforcement learning: An introduction},
	author={Sutton, Richard S and Barto, Andrew G},
	year={2018},
	publisher={MIT press}
}

@article{fremaux2013reinforcement,
	title={Reinforcement learning using a continuous time actor-critic framework with spiking neurons},
	author={Fr{\'e}maux, Nicolas and Sprekeler, Henning and Gerstner, Wulfram},
	journal={PLoS computational biology},
	volume={9},
	number={4},
	year={2013},
	publisher={Public Library of Science}
}

@article{florian2007reinforcement,
	title={Reinforcement learning through modulation of spike-timing-dependent synaptic plasticity},
	author={Florian, R{\u{a}}zvan V},
	journal={Neural Computation},
	volume={19},
	number={6},
	pages={1468--1502},
	year={2007},
	publisher={MIT Press}
}
\end{document}